%% file: main_arxiv.tex
\documentclass[10pt,twocolumn,letterpaper]{article}

\usepackage[pagenumbers]{cvpr} %


\definecolor{cvprblue}{rgb}{0.21,0.49,0.74}
\usepackage[pagebackref,breaklinks,colorlinks,allcolors=cvprblue]{hyperref}
\usepackage[accsupp]{axessibility}

\usepackage{xcolor}
\usepackage[table]{xcolor} 

\newcommand{\downpct}[2][red]{\textcolor{#1}{\ensuremath{\downarrow}\,#2\%}}
\newcommand{\uppct}[2][red]{\textcolor{#1}{\ensuremath{\uparrow}\,#2\%}}

\usepackage[numbers]{natbib}

\usepackage{bm}
\usepackage{color}
\usepackage{xcolor}
\usepackage{multirow}
\usepackage{ifthen}
\usepackage{dsfont}
\usepackage{pifont}%
\usepackage{booktabs}
\usepackage{graphicx}

\usepackage{nicefrac}
\usepackage{bigstrut}

\newcommand{\method}[1]{\ifthenelse{\equal{#1}{full}}{ColorTrigger}{ColorTrigger}}

\newcommand{\vparagraph}[1]{\noindent\textbf{#1}\quad}

\usepackage{pifont}%

\def\onedot{.}
\def\eg{\emph{e.g}\onedot}

 \def\vs{\emph{vs}\onedot}

\newcommand{\zs}[1]{\textcolor{black}{#1}}

\usepackage[normalem]{ulem}

\usepackage{algorithm}
\usepackage{algpseudocode} %

\def\paperID{2091} %
\def\confName{CVPR}
\def\confYear{2026}

\title{Color When It Counts: Grayscale-Guided Online Triggering \\
for Always-On Streaming Video Sensing}

\author{
Weitong Cai$^{1}$, 
Hang Zhang$^{2}$\footnotemark[1], 
Yukai Huang$^{3}$, 
Shitong Sun$^{1}$, \\
Jiankang Deng$^{4}$, 
Songcen Xu$^{5}$, 
Jifei Song$^{5}$, 
Zhensong Zhang$^{5}$\thanks{Corresponding authors.}\\
$^{1}$Queen Mary University of London, $^{2}$Independent Researcher, $^{3}$Durham University, \\
$^{4}$Imperial College London, $^{5}$Huawei Noah's Ark Lab\\
{\tt\small weitong.cai@qmul.ac.uk, miruku.hzhang@gmail.com, zhangzhensong@huawei.com}
}

\begin{document}
\maketitle
\input{sec/0_abstract}    
\input{sec/1_intro}
\input{sec/2_relatedwork}

\input{sec/3_method}
\input{sec/4_experiment}
\input{sec/5_conclusion}
\clearpage
{
    \small
    \bibliographystyle{ieeenat_fullname}
    \bibliography{main}
}

\input{sec/X_suppl}

\end{document}

%% file: sec/0_abstract.tex
\begin{abstract}
Always-on sensing is essential for next-generation edge/wearable AI systems, yet continuous high-fidelity RGB video capture remains prohibitively expensive for resource-constrained mobile and edge platforms. 
We present a new paradigm for efficient streaming video understanding: \textit{grayscale-always, color-on-demand}. 
Through preliminary studies, we discover that color is not always necessary. Sparse RGB frames suffice for comparable performance when temporal structure is preserved via continuous grayscale streams. 
Building on this insight, we propose \textbf{\method{}}, an online training-free trigger that selectively activates color capture based on windowed grayscale affinity analysis. 
Designed for real-time edge deployment, \method{} uses lightweight quadratic programming to detect chromatic redundancy causally, coupled with credit-budgeted control and dynamic token routing to jointly reduce sensing and inference costs. 
On streaming video understanding benchmarks, \method{} achieves \textbf{91.6\%} of full-color baseline performance while using only \textbf{8.1\%} RGB frames, demonstrating substantial color redundancy in natural videos and enabling practical always-on video sensing on resource-constrained devices. 
Project in: https://lvgd.github.io/ColorTrigger/
\end{abstract}

%% file: sec/1_intro.tex
\vspace{-5pt}
\section{Introduction}
\label{sec:intro}

\begin{figure}[t]
\centering
\includegraphics[width=1\linewidth]{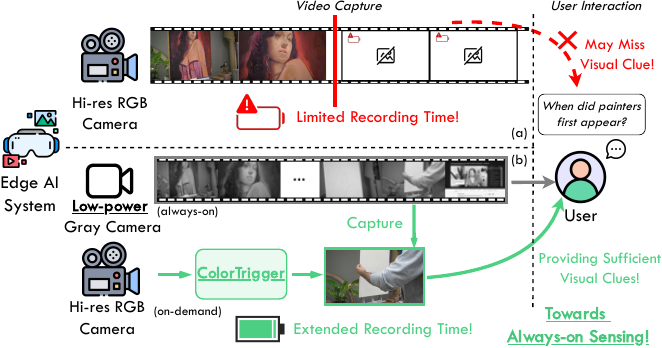}
\caption{
\textbf{Towards always-on sensing.}
(a) Realizing always-on sensing is difficult in practice: the Hi-res RGB video pipeline (such as sensor, ISP, encoding, and wireless transmission) quickly exhausts power on the edge AI system, so devices such as smart glasses typically sustain only about 30–60 minutes of continuous recording~\cite{meta2025rayban}, far from the all-day operation needed for a standby assistant.
(b) We propose \protect\method{}, a \emph{grayscale-always, color-on-demand} paradigm that uses a low-power gray camera as an always-on monitor and sparsely triggers an RGB camera only when needed, enabling always-on video sensing on edge devices.
}
\vspace{-15pt}
\label{fig:teaser} 
\end{figure}

\zs{Always-on sensing~\cite{paruchuri2025egotrigger,yang2025cambrian} is essential for building next-generation wearable and edge AI systems
that can seamlessly integrate perception and reasoning in real-time. Unlike traditional task-triggered systems that respond upon explicit user input, always-on AI continuously perceives and interprets the dynamic environment, enabling persistent understanding, memorizing, and modeling of the real world.}

However, \zs{realizing always-on sensing poses significant challenges in practice}, despite advances in multimodal large language models (MLLMs)~\cite{wang2024qwen2vl,zhang2024llavavideo,wang2025internvl35,cai2025mllm}, sustaining continuous high-fidelity vision capture over long horizons remains costly, especially for edge and wearable platforms (Figure~\ref{fig:teaser} (a)), \zs{\eg, a typical pair of smart glasses can only support about half an hour to one hour of continuous video recording~\cite{meta2025rayban}, making it infeasible to serve as an always-on, standby AI assistant}. 
Even when inference is offloaded, end-to-end energy bottlenecks are often dominated by continuous camera exposure
and wireless transport, while model-side cost scales with visual input density~\cite{paruchuri2025egotrigger}. 
\zs{Hence, achieving an effective balance among recording time, power consumption, bandwidth, and computational cost is critical for realizing always-on streaming video understanding. We argue that leveraging low-cost environmental sensing~\cite{himax_hm01b0} to selectively activate high-fidelity sensors~\cite{sony_imx681} offers a promising pathway toward this goal.}

\begin{figure}[t]
\centering
\begin{subfigure}{0.56\linewidth}
    \centering
    \includegraphics[width=\linewidth]{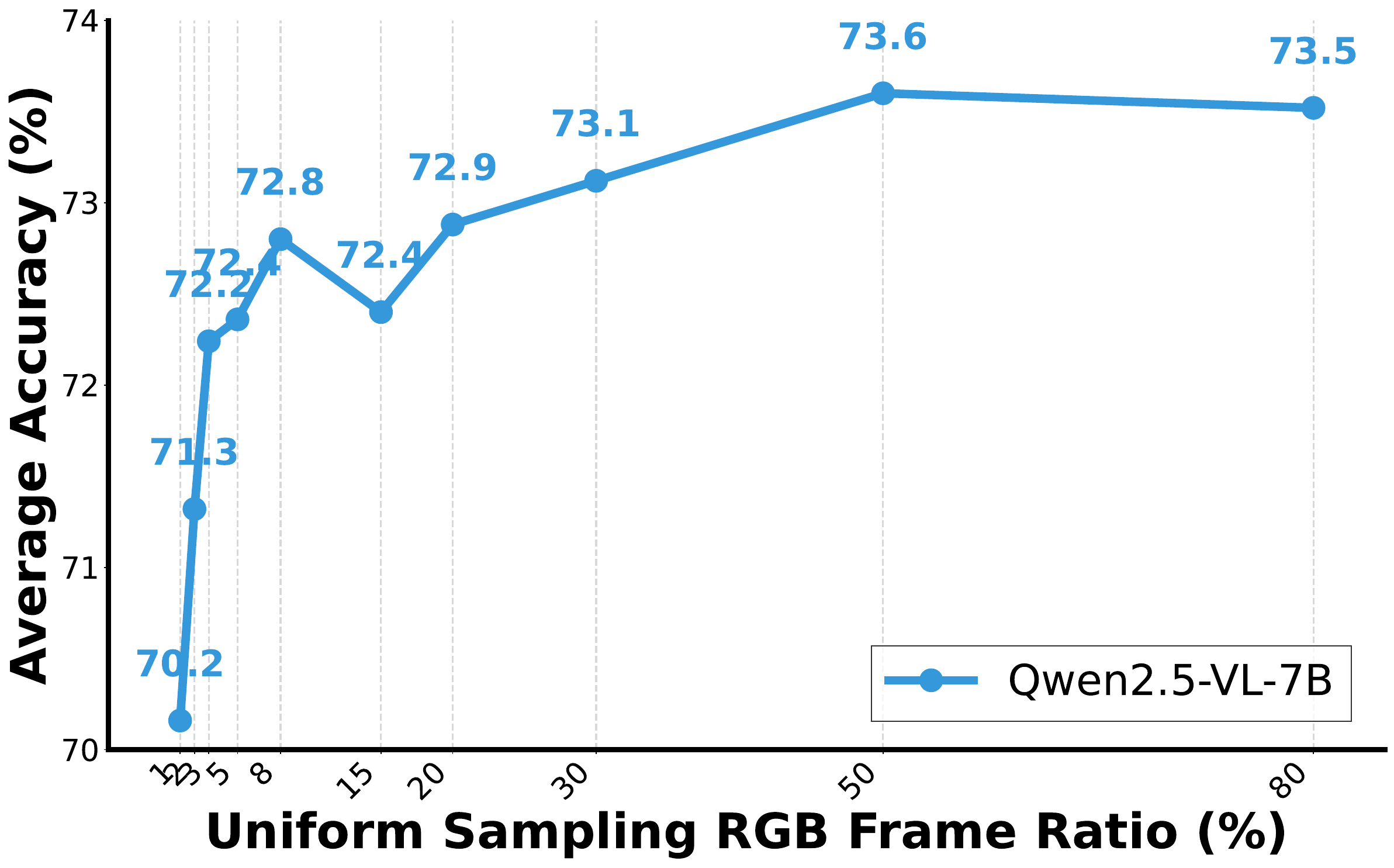}
    \caption{Performance vs. RGB ratio.}
    \label{fig:performance}
\end{subfigure}
\hfill
\begin{subfigure}{0.43\linewidth}
    \centering
    \includegraphics[width=\linewidth]{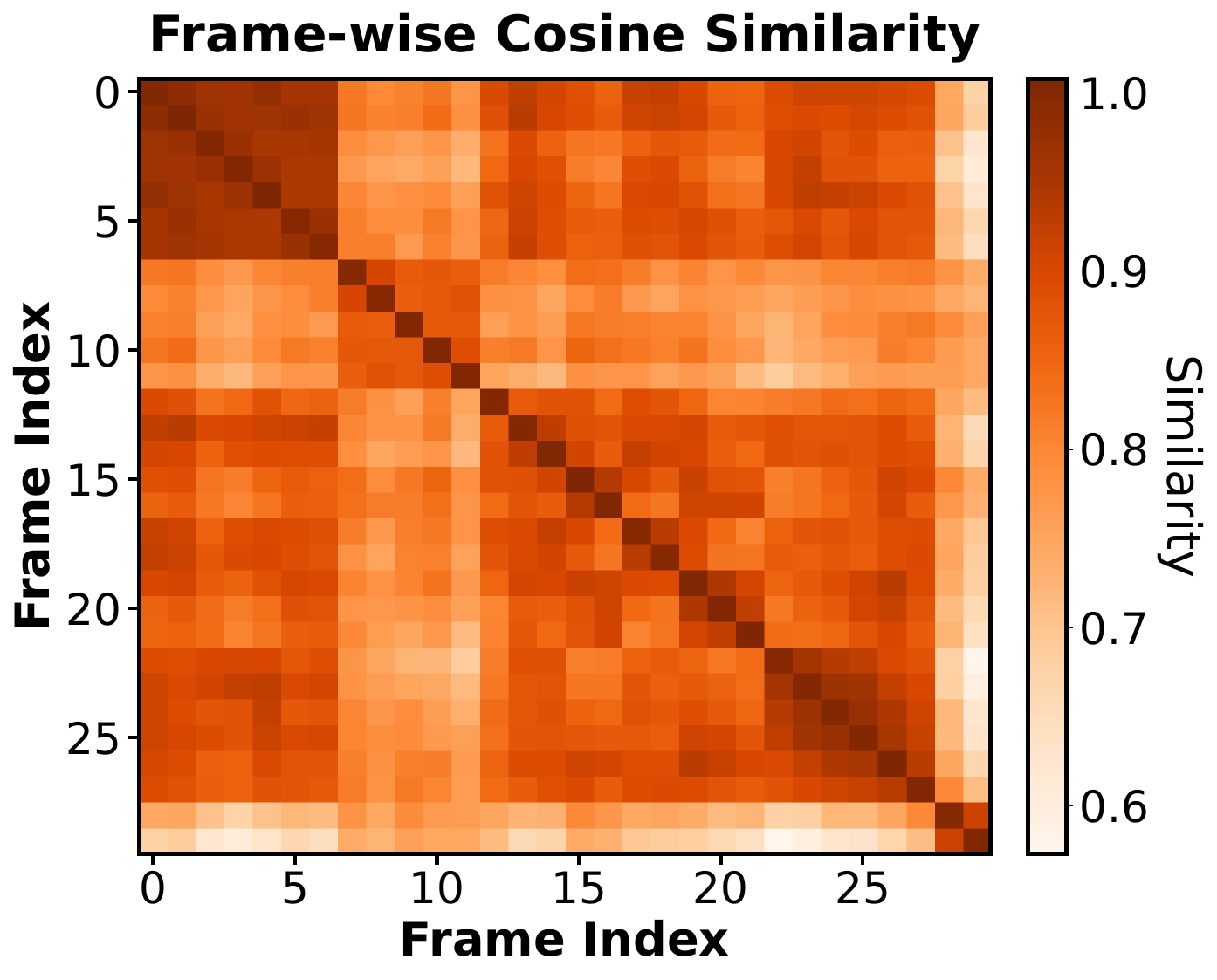}
    \caption{Temporal attention map.}
    \label{fig:attention}
\end{subfigure}
\caption{\textbf{Preliminary Studies: Color is not always necessary.}
(a) We uniformly insert RGB frames of the same resolution and quality into an otherwise grayscale video stream and evaluate on StreamingBench (All tasks) using Qwen2.5-VL-7B~\cite{bai2025qwen25vl}. Results show that only a small fraction of RGB frames is sufficient to achieve comparable performance. Furthermore,
(b) We extract CLS features from 30 consecutive frames of a 1 fps video using CLIP ViT-B/16~\cite{radford2021clip}, revealing redundancy in adjacent frames.
}
\vspace{-12pt}
\label{fig:combined}
\end{figure}

Existing approaches, such as EgoTrigger~\cite{paruchuri2025egotrigger}, reduce power consumption by triggering RGB cameras based on audio cues, but this creates prolonged periods with no visual information at all. When triggering fails, critical visual context is irretrievably lost with no fallback mechanism.
In our preliminary studies, \zs{we observe} that \textit{color is not always necessary} for video understanding with MLLMs.
Replacing a sparse set of RGB frames into an otherwise grayscale video stream yields steadily improved performance on video understanding benchmarks, as the RGB fraction increases, provided the temporal structure is preserved (Figure~\ref{fig:combined}). 
This phenomenon indicates substantial redundancy in the color dimension, suggesting that certain semantics, such as action recognition, layout reasoning, counting, and coarse recognition, are often largely color-independent, with chromatic detail benefiting only a subset of key moments/timestamps.

Motivated by this insight, we propose a \textit{grayscale-always, color-on-demand} paradigm for energy-efficient always-on streaming video understanding, using a low-power grayscale stream to monitor the scene and trigger RGB only when needed.
The grayscale stream runs continuously to preserve temporal structure and motion cues, while sparse color acquisitions are triggered only when necessary in a causal and online manner.
We introduce \textbf{\method{}}, a causal mechanism that treats color as an on-demand signal.
To enable real-time per-frame execution on resource-constrained edge devices, \method{} adopts a training-free design that requires no additional supervision.
A grayscale stream remains always on to preserve temporal continuity and minimize sensing cost, %
\zs{and a frozen visual encoder processes a sliding window to capture online redundancy and change.} 
We construct an online, windowed grayscale affinity matrix that summarizes recent redundancy and change.
Using only information from preceding frames, \method{} tries to infer whether the chromatic detail is increased at the current timestamp and selectively activate color capture while preserving causal operation.
Specifically, we cast the per-step decision as a lightweight continuous quadratic program (QP) that, using only the grayscale evidence in the current window, determines whether the present frame warrants color activation while remaining strictly causal.
To stabilize long-horizon behavior, a credit-budgeted online controller regulates when triggers may fire, ensuring the overall color spend remains bounded.
Given that, in practice, low-power grayscale cameras often deliver lower image quality and lower resolution than full-color modules, and to jointly reduce decoder-side compute, we further design \emph{dynamic token routing}: the grayscale stream traverses a high-compression pathway (fewer tokens), whereas triggered RGB frames follow a high-capacity pathway (more tokens). 
The resulting mixed sequence is reassembled in its native temporal order and fed directly to the downstream MLLM decoder without architectural changes or additional training. All trigger and routing decisions are derived solely from the grayscale stream, with no extra supervision, delivering color \textit{only when it counts} while preserving low latency and causal operation.

In summary, our contributions are threefold: 
\begin{itemize}
  \item \textbf{Finding: color is not always necessary.}
  We empirically characterize \emph{color redundancy} in video understanding.
  To our knowledge, this is among the first analyses of color redundancy for video understanding with MLLMs.

  \item \textbf{Grayscale-always, color-on-demand.} We introduce a practical paradigm for always-on streaming video sensing: a low-power grayscale stream preserves temporal structure, while a training-free online trigger selectively requests RGB via grayscale affinity analysis. The design integrates with dynamic token routing, maintaining causality while reducing sensing and inference cost.

  \item \textbf{Comparable performance under extreme RGB sparsity.}
  On streaming video understanding benchmarks,  \method{} achieves \textbf{91.6\%} of the full-color baseline performance while using only \textbf{8.1\%} RGB frames (\(91.9\%\) fewer than full RGB), indicating substantial natural redundancy of color information in streaming videos and validating the effectiveness of our proposed \method{}.
\end{itemize}

%% file: sec/2_relatedwork.tex
\begin{figure*}[t]
\centering
\includegraphics[width=1.0\textwidth]{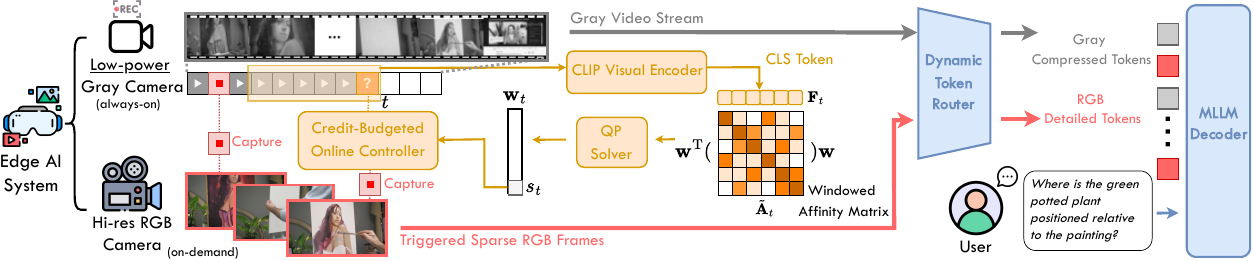}
\caption{
\textbf{Overview of \protect\method{}.} 
Our framework operates in two stages: (i) a causal online trigger analyzes grayscale features within a sliding window by aggregating them into an affinity matrix that captures temporal redundancy and change and applying a lightweight quadratic program under a credit-based budget; (ii) a dynamic token router adaptively allocates decoder capacity based on the trigger decision, using high-compression tokens for grayscale frames and high-capacity tokens for RGB frames, while preserving temporal order.
}
\label{fig:framework} 
\vspace{-10pt}
\end{figure*}

\section{Related Work}
\label{sec:formatting}

\vparagraph{Streaming video understanding with MLLMs.}
Recent efforts~\cite{chen2024videollmonline,di2025streaming,zhang2025flash,ning2025livevlm,cai2025mllm} adapt multimodal LLMs to \emph{online} video streams by reformulating training and inference for temporal alignment, long-context memory, and low-latency responses.
VideoLLM-online introduces LIVE to support real-time, temporally aligned dialogue over continuous streams~\cite{chen2024videollmonline}.
Follow-up work explores memory- and cache-centric streaming, such as ReKV which retrieves only query-relevant KV-caches for efficient StreamingVQA and enables training-free integration into existing Video-LLMs~\cite{di2025streaming}, and Flash-VStream which emphasizes real-time understanding on extremely long streams~\cite{zhang2025flash}.
Concurrently, LiveVLM studies long-context MLLM memory with rolling windows and KV compression~\cite{ning2025livevlm}.
\zs{Our method is complementary, instead of changing language or attention memory, we \emph{reduce RGB acquisition and processing} by grayscale-guided online triggering while keeping standard video-MLLM architectures.}

\vparagraph{Efficiency in MLLMs.}
\zs{A parallel line of work aims to reduce visual tokens and computation during inference via token pruning, routing, and dynamic resolution.} TokenLearner learns a small set of adaptive tokens for images/videos~\cite{ryoo2021tokenlearner}, ToMe merges redundant ViT tokens in a training-free manner~\cite{bolya2022tome}.
For MLLMs, ATP-LLaVA performs instance- and layer-wise \emph{adaptive} vision-token pruning with strong savings~\cite{ye2025atp}, and EfficientLLaVA auto-searches pruning policies with few samples~\cite{liang2025efficientllava}.
Qwen2-VL exposes a dynamic-resolution pathway that scales visual inputs pragmatically for efficiency~\cite{wang2024qwen2vl}, while InternVL3.5 introduces a Visual Resolution Router (ViR) for resolution-aware efficiency~\cite{wang2025internvl35}.
These methods operate \emph{after} frames are sampled; in contrast, we reduce cost partially \emph{before} tokenization by \textit{acquiring color only when needed}.

\vparagraph{Frame/keyframe selection and video summarization.}
Classical subset selection promotes importance and diversity via submodular objectives~\cite{gygli2015video} and determinantal point process (DPP)-based formulations~\cite{zhang2016summary,sharghi2018improving}.
Recent MLLM-oriented frame selectors begin to appear; e.g., BOLT selects informative frames at inference time for long videos~\cite{liu2025bolt}.
\zs{However, most prior work is offline and targets semantic coverage, whereas we focus on \emph{online, low-cost on-device acquisition} under practical resource constraints.}

\vparagraph{Event-driven sensing and on-demand capture.}
Event cameras exemplify how \emph{activity-driven} sensing can yield substantial efficiency gains in resource-constrained settings by sampling only when brightness changes occur~\cite{gallego2020event}. 
Recent audio-driven methods, such as EgoTrigger~\cite{paruchuri2025egotrigger} on smart glasses and Listen to Look~\cite{gao2020listen} for action recognition, use inexpensive audio cues to trigger or preview visual processing in long video streams and thereby reduce redundant frames. 
Rather than relying on auxiliary modalities, we use a low-power grayscale camera for always-on monitoring and sparsely wake up a higher-cost RGB camera.
\zs{Event-camera–based triggering requires additional hardware, whereas our method can run on commodity RGB sensors that already support low-power grayscale modes.}

%% file: sec/3_method.tex
\section{Method}
\label{sec:method}

We propose \textbf{\method{}}, a \emph{grayscale-always, color-on-demand} framework that treats color as an on-demand signal rather than a continuously sampled modality.
A low-power grayscale stream runs continuously to preserve temporal structure and motion cues, while RGB frames are requested only when grayscale evidence indicates chromatic detail is likely beneficial.
\method{} comprises two components:
(i) a \emph{causal online trigger} that analyzes a sliding window of grayscale features to detect redundancy and novelty, converting this analysis into binary color decisions under a long-horizon budget constraint; and
(ii) a \emph{dynamic token router} that allocates asymmetric decoder capacity, grayscale frames following a high-compression path while triggered RGB frames follow a high-capacity path, all while preserving native temporal order.
The pipeline is training-free, strictly causal, and integrates with frozen MLLMs, offering a practical solution for energy-constrained streaming video understanding toward always-on sensing.

\subsection{Problem Formulation}
\label{sec:problem}

Consider a live grayscale video stream $\mathcal{V} = \{g_t\}_{t=1}^{T}$ processed in a strictly causal manner.
At each timestep $t$, the system \emph{always} observes the low-cost grayscale frame $g_t$ from a continuously running monitor and may \emph{optionally} request a corresponding higher-cost RGB frame $c_t$ on demand.
The grayscale stream provides temporal continuity at minimal power, while RGB acquisition incurs additional sensing, transmission, and decoding costs.

\zs{A binary trigger $u_t \in \{0, 1\}$ determines whether to acquire $c_t$: $u_t=1$ requests RGB, while $u_t=0$ retains only grayscale.
The trigger operates subject to causality, meaning $u_t$  depends solely on past grayscale inputs $\{g_s\}_{s \leq t}$ and internal state, with no access to future frames.}

The trigger decision $u_t$ also governs token allocation: grayscale frames are encoded via a high-compression tokenizer $\psi_g(\cdot)$ producing $T_g$ tokens, while RGB frames use a high-capacity tokenizer $\psi_c(\cdot)$ producing $T_c$ tokens ($T_c > T_g$).
The resulting token sequence, assembled in temporal order, is fed to a frozen MLLM decoder to answer queries $Q$ at arbitrary times.
Our goal is to minimize RGB usage $\sum_{t=1}^T u_t$ while preserving video understanding accuracy, subject to strict causal and training-free constraints.

\subsection{Causal Online Trigger}
\label{sec:trigger}

The online trigger must satisfy three requirements: \emph{(a) causality}, operating frame-by-frame without future look-ahead; \emph{(b) training-free}, requiring no additional supervision for deployment on edge devices; and \emph{(c) budget awareness}, enforcing long-horizon RGB constraints while permitting local bursts around novel events.
We achieve this through a two-stage design: first, grayscale features within a sliding window are aggregated into an affinity matrix that captures recent redundancy; second, a lightweight quadratic program identifies the current frame's importance relative to this context, gated by a credit-based budget controller.

\vparagraph{Sliding window design.}
To balance temporal context with computational efficiency, we maintain a causal sliding window $\mathcal{W}_t = \{\max(1, t\!-\!W\!+\!1), \ldots, t\}$ of the most recent $W$ frames at each timestep $t$, where $n_t = |\mathcal{W}_t| \leq W$ is the effective window size.
This design ensures that: (i) affinity computation remains tractable with fixed complexity $\mathcal{O}(W^2)$ independent of video length $T$; (ii) the trigger focuses on \emph{recent} redundancy and change rather than distant history; and (iii) the system remains responsive to local dynamics while preserving strict causality.

\vparagraph{Grayscale feature extraction.}
At each timestep $t$, we extract features from all frames in the causal window $\mathcal{W}_t$ using \zs{a frozen CLIP visual encoder $\phi(\cdot) $~\cite{radford2021clip}}.
For each frame \zs{$g_i \in \mathcal{W}_t$}, we extract its global representation via the encoder's CLS token and $\ell_2$-normalize to obtain a unit-norm descriptor:
\begin{equation}
\mathbf{f}_i = \frac{\phi(g_i)}{\|\phi(g_i)\|_2} \in \mathbb{R}^d,
\label{eq:feature}
\end{equation}
where $d$ is the embedding dimension.
Descriptors are stacked row-wise into a matrix $\mathbf{F}_t = [\mathbf{f}_i]_{g_i \in \mathcal{W}_t} \in \mathbb{R}^{n_t \times d}$.

\vparagraph{Windowed grayscale affinity matrix.}
To summarize redundancy and temporal change within the recent grayscale stream, we construct a \emph{windowed affinity matrix} from the feature matrix $\mathbf{F}_t$.
Computing the Gram matrix $\mathbf{F}_t \mathbf{F}_t^\top$ yields pairwise cosine similarities (since features are $\ell_2$-normalized), with entries in $[-1, 1]$.
To ensure positive semi-definiteness (PSD) and numerical stability for subsequent quadratic optimization, we apply an affine transformation and enforce unit diagonal:
\begin{equation}
\tilde{\mathbf{A}}_t = \frac{1}{2}\big(\mathbf{F}_t \mathbf{F}_t^\top + \mathbf{I}_{n_t}\big) \in [0,1]^{n_t \times n_t},
\label{eq:affinity}
\end{equation}
where $\mathbf{I}_{n_t}$ is the identity matrix.
This transformation maps similarities to $[0,1]$ while preserving the inherent geometry: high $\tilde{A}_{ij}$ indicates frames $i$ and $j$ are redundant (similar), whereas low $\tilde{A}_{ij}$ suggests novelty or change.
The diagonal entries $\tilde{A}_{ii}\!=\!1$ reflect perfect self-similarity, and the matrix remains symmetric and PSD by construction.
Importantly, $\tilde{\mathbf{A}}_t$ is strictly causal, computed solely from frames within the current window $\mathcal{W}_t$.

\vparagraph{Quadratic program for diversity-driven selection.}
Given the affinity matrix $\tilde{\mathbf{A}}_t$, we seek to identify which frames in the window $\mathcal{W}_t$ are \emph{non-redundant} and thus informative.
We formulate this as a continuous quadratic program (QP) that assigns a selection weight $\mathbf{w}_t \in [0,1]^{n_t}$ to each frame in the window:
\begin{equation}
\mathbf{w}_t = \arg\min_{\mathbf{w} \in [0,1]^{n_t}} \lambda \, \mathbf{w}^\top \tilde{\mathbf{A}}_t \mathbf{w}
\quad \text{s.t.} \quad \mathbf{1}^\top \mathbf{w} = m_t,
\label{eq:qp}
\end{equation}
where $\lambda > 0$ is a scaling parameter and $m_t \in [0, n_t]$ is a pseudo-budget provided by the controller (described below).
The quadratic form $\mathbf{w}^\top \tilde{\mathbf{A}}_t \mathbf{w}$ penalizes allocating weight to mutually similar frames: if weights $w_i$ and $w_j$ are both large and frames $i,j$ are similar ($\tilde{A}_{ij}$ high), the objective increases.
Minimizing this term thus encourages spreading the fixed budget $m_t$ across frames that are \emph{temporally diverse}, naturally prioritizing frames with novel or distinctive content.
The quadratic form captures second-order pairwise 
correlations within the window, modeling the full 
inter-frame correlation structure rather than simply 
comparing each frame against history independently.

The continuous weights $\mathbf{w}_t$ can be interpreted as soft selection probabilities or importance scores.
Crucially, we focus solely on the weight of the \emph{current frame} (the last element of $\mathbf{w}_t$), denoted $s_t = (\mathbf{w}_t)_{n_t}$.
A high score $s_t$ indicates that frame $t$ contributes significantly to the window's diversity, suggesting it contains information not well-represented by recent history and thus warrants RGB acquisition.
This formulation is training-free, relying purely on geometric relationships in the grayscale feature space, and efficiently solvable via standard QP solvers~\cite{diamond2016cvxpy,stellato2020osqp}.

\vparagraph{Credit-budgeted online controller.}
While the QP provides a principled mechanism for assessing frame importance, unconstrained triggering could lead to unbounded RGB usage or pathological over-triggering during dynamic scenes.
To stabilize long-horizon behavior and enforce budget constraints, we introduce a \emph{credit-based online controller} that regulates when triggers may fire.

The controller maintains a scalar credit balance $b_t \in [0, C]$, where $C$ is a capacity cap.
Credits accumulate at a target rate $r > 0$ per frame (reflecting the desired RGB acquisition rate) and are consumed upon each RGB trigger.
The credit evolves causally according to:
\begin{equation}
b_{t+1} = \text{clip}\big(b_t - u_t + r, \, 0, \, C\big),
\label{eq:credit_update}
\end{equation}
where $\text{clip}(x, a_1, a_2) = \max(a_1, \min(x, a_2))$.
This update ensures credits cannot go negative or exceed the cap, 
preventing both budget underflow and unbounded accumulation.
The controller exposes a per-window pseudo-budget $m_t$ to the QP via a monotone 
mapping $\psi_{\text{budget}}: [0, C] \to [0, m_{\max}]$, where $m_{\max} \leq W$.
We use a simple clipping operation:
\begin{equation}
m_t = \psi_{\text{budget}}(b_t) = \min(n_t,\, \lfloor b_t \rfloor),
\label{eq:budget_map}
\end{equation}
which directly converts available credits to an integer budget within 
the current window size $n_t$.
Given the QP-derived score $s_t$, the binary trigger decision is:
\begin{equation}
u_t = \mathbb{I}\big[s_t \geq \theta \;\wedge\; b_t \geq 1\big],
\label{eq:trigger_decision}
\end{equation}
where $\theta$ is a threshold hyperparameter and $\mathbb{I}[\cdot]$ is the indicator function.
This gating mechanism ensures two conditions are met: the current frame must score highly relative to the window's diversity (geometric criterion), \emph{and} sufficient credits must be available (budget criterion).

Summing the credit update Eq.~\eqref{eq:credit_update} over a horizon $T$ yields $\sum_{t=1}^T u_t \leq rT + C$, ensuring total RGB usage is bounded by the target rate plus the initial capacity, thereby enforcing long-term budget compliance.
This design decouples the geometric reasoning (QP) from budget enforcement (credit gate), maintaining modularity and interpretability.

\subsection{Dynamic Token Router}
\label{sec:router}

We propose a dynamic token router that adapts input resolution according to the trigger decision: at each timestep $t$, untriggered grayscale frames are tokenized at lower resolution to reduce computation, while triggered RGB frames are tokenized at higher resolution to retain chromatic detail.
This design is motivated by practical mobile and wearable deployment scenarios, where always-on grayscale monitoring relies on low-power sensors whose outputs carry considerably less information than selectively captured RGB frames.
Applying a uniform token budget across such heterogeneous inputs is inherently inefficient; our router instead allocates representational capacity on demand, all while preserving native temporal ordering and requiring neither retraining nor architectural modifications to the backbone.

\vparagraph{Asymmetric capacity conditioned on $u_t$.}
Both tokenization strategies apply the same frozen visual encoder but differ in input resolution: grayscale frames are processed at lower resolution to yield $T_g$ tokens, while RGB frames use higher resolution to produce $T_c > T_g$ tokens.
Let $\psi_g(g_t) \in \mathbb{R}^{T_g \times d_v}$ and $\psi_c(c_t) \in \mathbb{R}^{T_c \times d_v}$ denote the resulting representations. The per-frame token block is:
\begin{equation}
\mathbf{Z}_t = (1-u_t)\,\psi_g(g_t) \oplus u_t\,\psi_c(c_t),
\label{eq:token_router}
\end{equation}
where $\oplus$ denotes concatenation.
Token blocks $\{\mathbf{Z}_t\}_{t=1}^T$ are concatenated in chronological order and processed under a standard causal attention mask by the frozen decoder.
The computational cost scales as $\sum_{t=1}^{T}\big[(1-u_t)T_g + u_t T_c\big]$, which remains strictly below the uniform cost $T \cdot T_c$ when RGB triggers are sparse ($\sum_{t=1}^T u_t \ll T$), concentrating representational capacity on informative moments without sacrificing long-range temporal context.

Algorithm~\ref{alg:colortrigger} summarizes the complete pipeline.

\begin{algorithm}[t]
\footnotesize
\caption{\protect\method{}}
\label{alg:colortrigger}
\begin{algorithmic}[1]
\Require Grayscale stream $\{g_t\}_{t=1}^{T}$, window size $W$, threshold $\theta$, target rate $r$, credit cap $C$, encoder $\phi(\cdot)$
\Ensure Trigger sequence $\{u_t\}_{t=1}^{T}$ and token sequence $\{\mathbf{Z}_t\}_{t=1}^{T}$
\State Initialize credit balance $b_1 \gets C$
\For{$t = 1, 2, \ldots, T$}
    \State Extract features $\mathbf{F}_t$ from causal window $\mathcal{W}_t$ \Comment{Eq.~\eqref{eq:feature}}
    \State Construct affinity $\tilde{\mathbf{A}}_t \gets \frac{1}{2}(\mathbf{F}_t \mathbf{F}_t^\top + \mathbf{I})$ \Comment{Eq.~\eqref{eq:affinity}}
    \State Compute pseudo-budget $m_t \gets \psi_{\text{budget}}(b_t)$ \Comment{Eq.~\eqref{eq:budget_map}}
    \State Solve diversity QP: $\mathbf{w}_t \gets \arg\min \lambda \mathbf{w}^\top \tilde{\mathbf{A}}_t \mathbf{w}$ s.t. $\mathbf{1}^\top \mathbf{w} = m_t$ \Comment{Eq.~\eqref{eq:qp}}
    \State Compute importance score $s_t \gets (\mathbf{w}_t)_{n_t}$ for current frame
    \State Trigger decision: $u_t \gets \mathbb{I}[s_t \ge \theta \;\wedge\; b_t \ge 1]$ \Comment{$b_t$ updated by Eq.~\eqref{eq:credit_update}}
    \State Route tokens: $\mathbf{Z}_t \gets (1-u_t)\,\psi_g(g_t) \oplus u_t\,\psi_c(c_t)$ \Comment{Eq.~\eqref{eq:token_router}}
    \State Feed $\mathbf{Z}_t$ to frozen MLLM decoder
    \State Update credit: $b_{t+1} \gets \mathrm{clip}(b_t - u_t + r,\, 0,\, C)$ \Comment{Eq.~\eqref{eq:credit_update}}
\EndFor
\end{algorithmic}
\end{algorithm}

%% file: sec/4_experiment.tex
\section{Experiment}
\label{sec:experiment}

\begin{table*}[tb]
\centering
\caption{Performance comparison on StreamingBench~\cite{lin2024streamingbench} focusing on \textit{Real-Time Visual Understanding} tasks. Real-Time Visual Understanding encompasses Object Perception (OP), Causal Reasoning (CR), Clips Summarization (CS), Attribute Perception (ATP), Event Understanding (EU), Text-Rich Understanding (TR), Prospective Reasoning (PR), Spatial Understanding (SU), Action Perception (ACP), and Counting (CT). ``{RGB(\%)}'' represents the percentage of color frames remaining after triggering, where $100$ indicates full color, $0$ means all grayscale, and $\downarrow91.9\%$ signifies a 91.9\% reduction in color frames. 
}
\resizebox{\linewidth}{!}{%
\input{table/streamingbench}
}
\label{tab:streamingbench}
\end{table*}

\begin{table*}[tb]
\centering
\caption{Results on OVO-Bench~\cite{niu2025ovo} comprising three categories: i) \textit{Real-Time Visual Perception} (OCR: Optical Character Recognition, ACR: Action Recognition, ATR: Attribute Recognition, STU: Spatial Understanding, FPD: Future Prediction, OJR: Object Recognition), ii) \textit{Backward Tracing} (EPM: Episodic Memory, ASI: Action Sequence Identification, HLD: Hallucination Detection), and iii) \textit{Forward Active Responding} (REC: Repetition Event Count, SSR: Sequential Steps Recognition, CRR: Clues Reveal Responding).}    
\resizebox{\linewidth}{!}{%
\input{table/ovobench.tex}
}
\label{tab:ovobench}
\end{table*}

\begin{table*}[th]  %
\begin{minipage}[t]{0.62\linewidth}
\centering
\caption{Performance comparison on Video-MME~\cite{fu2025videomme}.}
\resizebox{\linewidth}{!}{%
\input{table/videomme}
}
\label{tab:videomme}
\end{minipage}
\hfill
\begin{minipage}[t]{0.37\linewidth}
\footnotesize
\centering
\caption{Component Study. $\Delta$ indicates the relative performance (\%) w.r.t. baseline. UniRGB and TrigRGB represent uniformly sampled and triggered RGB frames, respectively.}
\renewcommand{\arraystretch}{1.015}  %
\resizebox{\linewidth}{!}{%
\input{table/component.tex}
}
\label{tab:component}
\end{minipage}
\end{table*}

\vparagraph{Implementation details.}
We use InternVL3.5-8B-Instruct~\cite{wang2025internvl35} as the frozen MLLM backbone. 
For the causal trigger, we employ CLIP ViT-B/16~\cite{radford2021clip} as the frozen feature extractor %
due to its computational efficiency and strong transferability.
The grayscale stream captures only the L channel (luminance) in the CIELAB color space following~\cite{kang2023ddcolor}.
Grayscale frames are resized to $224 \times 224$ before patchification, yielding $T_g = 64$ visual tokens after spatial compression, 
while RGB frames are processed at $448 \times 448$, producing $T_c = 256$ tokens.
For the trigger, we set window size $W=30$, threshold $\theta=0.3$, target RGB rate $r \in \{0.1, 0.4\}$, and QP scaling parameter $\lambda=1.0$.
The credit capacity is set to $C = 2$.
To avoid overly conservative triggering in dynamic scenes, we augment the theoretical budget $\lfloor b_t \rfloor$ with a small lookahead buffer in practice:
$m_t = \min(n_t,\, \lfloor b_t + r \cdot L \rfloor),$
where $L=10$ allows the controller to anticipate near-future credit accumulation, 
smoothing short-term fluctuations. 
We solve the QP using CVXPY~\cite{diamond2016cvxpy} with the OSQP solver~\cite{stellato2020osqp}. 
We sample frames at 1 FPS with a maximum of 128 frames per video.

\subsection{Results on Streaming Video Benchmarks}
We evaluate our model on two popular streaming VideoQA benchmarks: StreamingBench~\cite{lin2024streamingbench} and OVO-Bench~\cite{niu2025ovo}. Following the streaming setting, \method{} processes all historical frames accumulated up to the \textit{current} timestamp when each question is posed, enabling real-time response without access to future content.

\vparagraph{StreamingBench.} 
Table~\ref{tab:streamingbench} demonstrates that \method{} achieves competitive performance on the Real-time Visual Understanding subtask of StreamingBench. Our model with 34.3\% RGB frames scores 75.24, outperforming recent online model Dispider-7B and close to TimeChat-Online-7B, while being comparable to proprietary models such as Gemini 1.5 Pro (75.69) and surpassing GPT-4o (73.28) and Claude 3.5 Sonnet (72.44).
Compared to the base model InternVL-3.5-8B with 100\% RGB frames (77.20), \method{} achieves 75.24 while reducing RGB frame usage by 65.7\%, demonstrating an effective trade-off between performance and computational efficiency. 
Notably, even with only 8.1\% RGB frames (91.9\% reduction), our approach scores 70.72, showing an 8.64\% improvement over the grayscale-only baseline (62.08) and remaining competitive with many existing streaming models. 

\noindent\textbf{OVO-Bench.} 
Table~\ref{tab:ovobench} presents results on OVO-Bench, which comprehensively evaluates streaming video understanding across three categories: Real-Time Visual Perception, Backward Tracing, and Forward Active Responding. Our model with 33.1\% RGB frames achieves an overall score of 52.5, outperforming almost all existing open-source online MLLMs.
Compared to the base model InternVL-3.5-8B with full RGB input (57.7), \method{} scores 52.5 while reducing RGB frame usage by 66.9\%, representing only a 5.2-point drop in overall performance. This modest degradation is accompanied by substantial gains in efficiency, demonstrating the effectiveness of our adaptive routing strategy. Notably, our Real-Time Visual Perception performance (65.2) shows an 11.4-point improvement over the grayscale-only baseline (53.8), highlighting the importance of selectively introducing chromatic information at critical moments. Even with only 7.1\% RGB frames (92.9\% reduction), \method{} maintains a competitive overall score of 50.4, a 2.5 improvement over the grayscale-only setting.

These results highlight two key findings: (1) substantial redundancy exists in full RGB video streams for streaming understanding tasks, and (2) our adaptive routing mechanism effectively identifies critical moments requiring chromatic information, enabling strong performance with minimal RGB frame acquisition.

\begin{figure*}[th]
\centering
\includegraphics[width=1\textwidth]{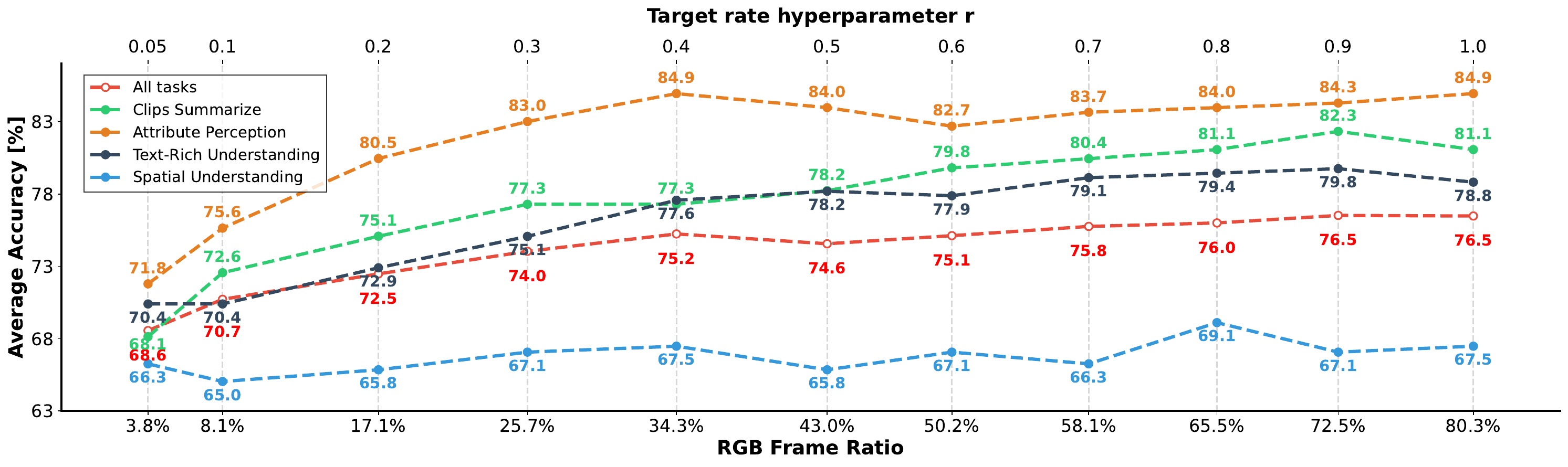}

\caption{
\textbf{Performance across varying RGB frame ratios.} 
We vary the target rate $r \in [0.05, 1.0]$ and evaluate on StreamingBench.
}
\label{fig:perratio} 
\end{figure*}

\subsection{Results on Offline Long Video Task}
Table~\ref{tab:videomme} presents results on Video-MME~\cite{fu2025videomme}, a comprehensive long-form video understanding benchmark spanning short, medium, and long duration videos. Our model with 37.6\% RGB frames achieves an overall score of 66.1, surpassing the full RGB baseline InternVL-3.5-8B at 65.6 while using 62.4\% fewer chromatic frames. This demonstrates that our adaptive triggering mechanism not only reduces computational cost but can actually improve performance by focusing RGB capacity on semantically critical moments. Notably, \method{} outperforms all existing online MLLMs including TimeChat-Online-7B at 62.4 and Dispider-7B at 57.2, confirming the effectiveness of combining continuous grayscale context with selective RGB acquisition for long-form video understanding. Even with only 9.1\% RGB frames, our model scores 62.8, representing a 5.5-point improvement over the grayscale-only baseline at 57.3 and remaining competitive with many full RGB methods. The consistent performance gains across all duration categories validate that our approach maintains temporal coherence and captures critical visual information effectively across varying video lengths.

\subsection{Ablation Study}
\vparagraph{Component Study.}
Table~\ref{tab:component} validates our design choices on StreamingBench. At 8.1\% RGB usage, our full model (\textit{Grayscale + TrigRGB}) achieves 70.72, recovering 91.6\% of baseline performance. Removing continuous grayscale (\textit{Only TrigRGB}: 68.76) causes a 1.96-point drop, while replacing adaptive triggering with uniform sampling results in a 1.72-point degradation, demonstrating that both components are essential. At 34.3\% RGB usage, our model scores 75.24 (97.5\% recovery), consistently outperforming uniform sampling by 2.60 points and RGB-only configuration by 1.76 points. These results confirm that continuous grayscale provides crucial temporal context and adaptive triggering effectively identifies semantically important moments, with their combination enabling strong performance at minimal RGB usage.

\vparagraph{Performance across varying RGB frame ratios.}
Figure~\ref{fig:perratio} examines the impact of target rate $r$ on RGB usage and task-specific performance. 
As $r$ increases from 0.05 to 1.0, the RGB frame ratio scales proportionally from 3.8\% to 80.3\%, validating our credit-based controller's effectiveness in regulating color acquisition. 
Overall accuracy increases monotonically, confirming that \emph{color is beneficial but not always necessary}. 
Task-specific trends reveal heterogeneous color dependencies: Attribute Perception rapidly saturates at ${\sim}26\%$ RGB ratio, demonstrating substantial chromatic redundancy where sparse color frames suffice; Clips Summarization and Text-Rich Understanding exhibit gradual gains, reflecting dependencies on temporal coherence and fine-grained detail; Spatial Understanding remains largely flat (65.0--69.1\%), indicating that reasoning about spatial relations requires minimal chromatic information and can be adequately addressed with grayscale alone, further evidence of color redundancy. 
Importantly, \method{} operates question-agnostically with purely vision-based triggering without query-specific optimization, thus the QA performance fluctuations across adjacent RGB ratios are expected under our design.

\begin{table}[t]
\centering
\caption{Token Efficiency Analysis. $\Delta$ denotes the percentage relative to baseline. DT represents Dynamic Token Router, and TrigRGB represents triggered RGB frame selection.}
\resizebox{\linewidth}{!}{%
\input{table/tokenreduce.tex}
}
\label{tab:tokenreduce}
\end{table}

\vparagraph{Token Efficiency Analysis.}
Table~\ref{tab:tokenreduce} demonstrates the effectiveness of our Dynamic Token Router (DT) in reducing computational cost. Without dynamic tokenization, maintaining 256 grayscale tokens per frame with 8.1\% RGB usage achieves 73.68 but consumes 100\% of baseline visual tokens. By introducing DT with 64 tokens for grayscale frames, our model reduces total visual tokens to only 31.1\% while scoring 70.72 (91.6\% performance recovery), representing a favorable trade-off between efficiency and accuracy. At 34.3\% RGB usage, DT achieves 75.24 with 50.7\% token consumption, closely matching the 256-token variant (76.32) that uses twice the computational resources. 
These results validate that our asymmetric tokenization strategy effectively allocates capacity, concentrating tokens on high-fidelity RGB frames while maintaining temporal coherence through compressed grayscale representations, enabling substantial computational savings without proportional performance degradation.

\begin{figure}[t]
\centering
\includegraphics[width=1\linewidth]{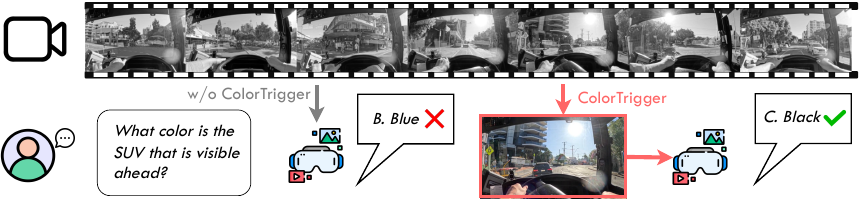}
\caption{
Qualitative Example of the proposed method.
}
\label{fig:visualization} 
\end{figure}

\vparagraph{Visualization.}
We provide an illustrative example in Figure~\ref{fig:visualization}, where the MLLM processes only grayscale video streams and fails to answer ``What color is the SUV?", incorrectly predicting ``Blue". In contrast, our \method{} effectively triggers and captures the relevant frames, obtaining high-resolution RGB images. With access to color information, the model correctly identifies the answer as ``Black", demonstrating the necessity and efficacy of our adaptive RGB frame triggering mechanism.

Feasibility analyses, additional experiments on 
Qwen3-VL~\cite{yang2025qwen3}, qualitative 
visualizations, and discussion of limitations are 
provided in the supplementary material.

%% file: table/streamingbench.tex
\begin{tabular}{lccccccccccccc}
\hline
Model & \#Frames & RGB (\%) & OP    & CR    & CS    & ATP   & EU    & TR    & PR    & SU    & ACP   & CT    & All \bigstrut\\
\hline
\hline
Human & -     & -     & 89.47 & 92.00 & 93.60 & 91.47 & 95.65 & 92.52 & 88.00 & 88.75 & 89.74 & 91.30 & 91.46 \bigstrut\\
\hline
\multicolumn{14}{c}{\textbf{Proprietary MLLMs}} \bigstrut\\
\hline
Gemini 1.5 Pro~\cite{team2024gemini} & 1 fps & 100   & 79.02 & 80.47 & 83.54 & 79.67 & 80.00 & 84.74 & 77.78 & 64.23 & 71.95 & 48.70 & 75.69 \bigstrut[t]\\
GPT-4o~\cite{hurst2024gpt} & 64    & 100   & 77.11 & 80.47 & 83.91 & 76.47 & 70.19 & 83.80 & 66.67 & 62.19 & 69.12 & 49.22 & 73.28 \\
Claude 3.5 Sonnet~\cite{claude3_5} & 20    & 100   & 80.49 & 77.34 & 82.02 & 81.73 & 72.33 & 75.39 & 61.11 & 61.79 & 69.32 & 43.09 & 72.44 \bigstrut[b]\\
\hline
\multicolumn{14}{c}{\textbf{Open-Source Video MLLMs }} \bigstrut\\
\hline
LLaVA-OneVision-7B~\cite{li2024llavaonevision} & 32    & 100   & 80.38 & 74.22 & 76.03 & 80.72 & 72.67 & 71.65 & 67.59 & 65.45 & 65.72 & 45.08 & 71.12 \bigstrut[t]\\
Video-LLaMA2-7B~\cite{cheng2024videollama2} & 32    & 100   & 55.86 & 55.47 & 57.41 & 58.17 & 52.80 & 43.61 & 39.81 & 42.68 & 45.61 & 35.23 & 49.52 \\
Qwen2.5-VL-7B~\cite{bai2025qwen25vl} & 1 fps & 100   & 78.32 & 80.47 & 78.86 & 80.45 & 76.73 & 78.50 & 79.63 & 63.41 & 66.19 & 53.19 & 73.68 \bigstrut[b]\\
\hline
\multicolumn{14}{c}{\textbf{Streaming MLLMs}} \bigstrut\\
\hline
Flash-VStream-7B~\cite{zhang2024flash} & -     & 100   & 25.89 & 43.57 & 24.91 & 23.87 & 27.33 & 13.08 & 18.52 & 25.20 & 23.87 & 48.70 & 23.23 \bigstrut[t]\\
VideoLLM-online-8B~\cite{chen2024videollmonline} & 2 fps & 100   & 39.07 & 40.06 & 34.49 & 31.05 & 45.96 & 32.40 & 31.48 & 34.16 & 42.49 & 27.89 & 35.99 \\
Dispider-7B~\cite{qian2025dispider} & 1 fps & 100   & 74.92 & 75.53 & 74.10 & 73.08 & 74.44 & 59.92 & 76.14 & 62.91 & 62.16 & 45.80 & 67.63 \\
TimeChat-Online-7B~\cite{yao2025timechatonline} & 1 fps & 100   & 80.22 & 82.03 & 79.50 & 83.33 & 76.10 & 78.50 & 78.70 & 64.63 & 69.60 & 57.98 & 75.36 \bigstrut[b]\\
\hline
InternVL-3.5-8B & 128   & 100   & 83.47 & 82.03 & 82.65 & 84.62 & 75.47 & 80.06 & 81.48 & 67.89 & 70.45 & 59.04 & 77.20 \bigstrut[t]\\
\rowcolor[rgb]{ .91,  .91,  .91} InternVL-3.5-8B grayscale & 128   & 0     & 60.98 & 78.91 & 60.88 & 55.45 & 65.41 & 63.24 & 75.00 & 60.16 & 62.22 & 55.85 & 62.08 \\
\rowcolor[rgb]{ .984,  .886,  .835} \method{} (Ours, 8B) & 128   & 8.1 (\downpct{91.9}) & 75.07 & 77.34 & 72.56 & 75.64 & 71.70 & 70.40 & 76.85 & 65.04 & 66.48 & 57.98 & 70.72 (\uppct{8.64}) \\
\rowcolor[rgb]{ .969,  .78,  .675} \method{} (Ours, 8B) & 128   & 34.3 (\downpct{65.7}) & 81.57 & 78.91 & 77.29 & 84.94 & 76.10 & 77.57 & 81.48 & 67.48 & 67.61 & 56.91 & 75.24 (\uppct{13.16}) \bigstrut[b]\\
\hline
\end{tabular}%

%% file: table/ovobench.tex
\begin{tabular}{lcccccccccccccccccc}
\hline
\multirow{2}[4]{*}{Model} & \multirow{2}[4]{*}{\#Frames} & \multirow{2}[4]{*}{RGB (\%)} & \multicolumn{7}{c}{Real-Time Visual Perception}       & \multicolumn{4}{c}{Backward Tracing} & \multicolumn{4}{c}{Forward Active Responding} & Overall \bigstrut\\
\cline{4-18}      &       &       & OCR   & ACR   & ATR   & STU   & FPD   & OJR   & Avg.  & EPM   & ASI   & HLD   & Avg.  & REC   & SSR   & CRR   & Avg.  & Avg. \bigstrut\\
\hline
\hline
Human Agents & -     & -     & 94.0  & 92.6  & 94.8  & 92.7  & 91.1  & 94.0  & 93.2  & 92.6  & 93.0  & 91.4  & 92.3  & 95.5  & 89.7  & 93.6  & 92.9  & 92.8 \bigstrut\\
\hline
\multicolumn{19}{c}{\textbf{Proprietary MLLMs}} \bigstrut\\
\hline
Gemini 1.5 Pro~\cite{team2024gemini} & 1 fps & 100   & 87.3  & 67.0  & 80.2  & 54.5  & 68.3  & 67.4  & 70.8  & 68.6  & 75.7  & 52.7  & 62.3  & 35.5  & 74.2  & 61.7  & 57.2  & 65.3 \bigstrut[t]\\
GPT-4o~\cite{hurst2024gpt} & 64    & 100   & 69.1  & 65.1  & 65.5  & 50.0  & 68.3  & 63.7  & 63.6  & 49.8  & 71.0  & 55.4  & 58.7  & 27.6  & 73.2  & 59.4  & 53.4  & 58.6 \bigstrut[b]\\
\hline
\multicolumn{19}{c}{\textbf{Open-Source Video MLLMs }} \bigstrut\\
\hline
LLaVA-NeXT-Video-7B~\cite{llavanextvideo} & 64    & 100   & 69.8  & 59.6  & 66.4  & 50.6  & 72.3  & 61.4  & 63.3  & 51.2  & 64.2  & 9.7   & 41.7  & 34.1  & 67.6  & 60.8  & 54.2  & 53.1 \bigstrut[t]\\
LLaVA-OneVision-7B~\cite{li2024llavaonevision} & 64    & 100   & 67.1  & 58.7  & 69.8  & 49.4  & 71.3  & 60.3  & 62.8  & 52.5  & 58.8  & 23.7  & 45.0  & 24.8  & 66.9  & 60.8  & 50.9  & 52.9 \\
Qwen2-VL-7B~\cite{wang2024qwen2vl} & 64    & 100   & 69.1  & 53.2  & 63.8  & 50.6  & 66.3  & 60.9  & 60.7  & 44.4  & 66.9  & 34.4  & 48.6  & 30.1  & 65.7  & 50.8  & 48.9  & 52.7 \\
LongVU-7B~\cite{shen2024longvu} & 1 fps & 100   & 55.7  & 49.5  & 59.5  & 48.3  & 68.3  & 63.0  & 57.4  & 43.1  & 66.2  & 9.1   & 39.5  & 16.6  & 69.0  & 60.0  & 48.5  & 48.5 \bigstrut[b]\\
\hline
\multicolumn{19}{c}{\textbf{Streaming MLLMs}} \bigstrut\\
\hline
Flash-VStream-7B~\cite{zhang2024flash} & 1 fps & 100   & 25.5  & 32.1  & 29.3  & 33.7  & 29.7  & 28.8  & 29.9  & 36.4  & 33.8  & 5.9   & 25.4  & 5.4   & 67.3  & 60.0  & 44.2  & 33.2 \bigstrut[t]\\
VideoLLM-online-8B~\cite{chen2024videollmonline} & 2 fps & 100   & 8.1   & 23.9  & 12.1  & 14.0  & 45.5  & 21.2  & 20.8  & 22.2  & 18.8  & 12.2  & 17.7  & -     & -     & -     & -     & - \\
TimeChat-Online-7B~\cite{yao2025timechatonline} & 1 fps & 100   & 75.2  & 46.8  & 70.7  & 47.8  & 69.3  & 61.4  & 61.9  & 55.9  & 59.5  & 9.7   & 41.7  & 31.6  & 38.5  & 40.0  & 36.7  & 46.7 \\
Qwen2.5-VL-7B~\cite{bai2025qwen25vl} & 1 fps & 100   & 73.8  & 56.0  & 68.1  & 46.6  & 71.3  & 60.3  & 62.7  & 48.2  & 64.9  & 26.9  & 46.6  & 36.2  & 41.8  & 47.1  & 41.7  & 50.3 \bigstrut[b]\\
\hline
InternVL-3.5-8B & 128   & 100   & 77.9  & 64.2  & 75.9  & 56.7  & 75.3  & 70.7  & 70.1  & 56.6  & 66.2  & 34.4  & 52.4  & 48.2  & 62.7  & 41.3  & 50.7  & 57.7 \bigstrut[t]\\
\rowcolor[rgb]{ .91,  .91,  .91} InternVL-3.5-8B grayscale & 128   & 0     & 61.1  & 46.8  & 53.5  & 47.8  & 56.4  & 57.1  & 53.8  & 46.1  & 54.7  & 37.1  & 46.0  & 35.3  & 54.3  & 42.1  & 43.9  & 47.9 \\
\rowcolor[rgb]{ .984,  .886,  .835} \method{} (Ours, 8B) & 128   & 7.1 (\downpct{92.9}) & 65.1  & 52.3  & 61.2  & 50.6  & 66.3  & 63.6  & 59.9 (\uppct{6.1}) & 50.5  & 58.1  & 32.3  & 47.0  & 32.5  & 59.3  & 41.3  & 44.3  & 50.4 (\uppct{2.5}) \\
\rowcolor[rgb]{ .969,  .78,  .675} \method{} (Ours, 8B) & 128   & 33.1  (\downpct{66.9}) & 77.9  & 56.0  & 69.8  & 55.1  & 67.3  & 65.2  & 65.2 (\uppct{11.4}) & 50.2  & 62.8  & 30.7  & 41.3  & 32.7  & 59.4  & 40.8  & 44.3  & 52.5 (\uppct{4.6}) \bigstrut[b]\\
\hline
\end{tabular}%

%% file: table/videomme.tex
\begin{tabular}{lcccccc}
\hline
\multirow{2}[4]{*}{Method} & \multirow{2}[4]{*}{\#Frames} & \multirow{2}[4]{*}{RGB (\%)} & \multicolumn{4}{c}{Video-MME} \bigstrut\\
\cline{4-7}      &       &       & short & medium & long  & overall \bigstrut\\
\hline
\hline
VideoChat2-7B~\cite{li2024mvbench} & 16    & 100   & 48.3  & 37.0  & 33.2  & 39.5 \bigstrut[t]\\
LongVA-7B~\cite{zhang2024longva} & 128   & 100   & 61.1  & 50.4  & 46.2  & 52.6 \\
Kangaroo-7B~\cite{kangaroogroup} & 64    & 100   & 66.1  & 55.3  & 46.6  & 56.0 \\
Video-CCAM-14B~\cite{fei2024video} & 96    & 100   & 62.2  & 50.6  & 46.7  & 53.2 \\
VideoXL-7B~\cite{shu2025videoxl} & 128   & 100   & 64.0  & 53.2  & 49.2  & 55.5 \\
Dispider-7B~\cite{qian2025dispider} & 1 fps & 100   & -     & -     & -     & 57.2 \\
VideoChat-Online-4B~\cite{huang2024online} & 2 fps & 100   & -     & -     & 47.1  & 54.4 \\
TimeChat-Online-7B~\cite{yao2025timechatonline} & 1 fps & 100   & -     & -     & 48.4  & 62.4 \bigstrut[b]\\
\hline
InternVL-3.5-8B & 128   & 100   & 76.7  & 65.3  & 54.7  & 65.6 \bigstrut[t]\\
\rowcolor[rgb]{ .91,  .91,  .91} InternVL-3.5-8B grayscale & 128   & 0     & 63.4  & 57.9  & 50.6  & 57.3 \\
\rowcolor[rgb]{ .984,  .886,  .835} \method{} (Ours, 8B) & 128   & 9.1 (\downpct{90.9}) & 71.9  & 63.0  & 53.6  & 62.8 (\uppct{5.5}) \\
\rowcolor[rgb]{ .969,  .78,  .675} \method{} (Ours, 8B) & 128   & 37.6 (\downpct{62.4}) & 76.8  & 66.3  & 55.1  & 66.1 (\uppct{8.8}) \bigstrut[b]\\
\hline
\end{tabular}%

%% file: table/component.tex
\begin{tabular}{lccc}
\hline
Model & RGB (\%) & All   & $\Delta$ \bigstrut\\
\hline
\hline
Baseline & 100   & 77.20 & 100.0\% \bigstrut[t]\\
Grayscale & 0     & 62.08 & 80.4\% \bigstrut[b]\\
\hline
Only UniRGB & 8.1   & 65.52 & 84.9\% \bigstrut[t]\\
Grayscale + UniRGB & 8.1   & 69.00 & 89.4\% \\
Only TrigRGB & 8.1   & 68.76 & 89.1\% \\
\rowcolor[rgb]{ .984,  .886,  .835} Grayscale + TrigRGB (Ours) & 8.1   & \textbf{70.72} & \textbf{91.6\%} \bigstrut[b]\\
\hline
Only UniRGB & 34.3  & 70.84 & 91.8\% \bigstrut[t]\\
Grayscale + UniRGB & 34.3  & 72.64 & 94.1\% \\
Only TrigRGB & 34.3  & 73.48 & 95.2\% \\
\rowcolor[rgb]{ .984,  .886,  .835} Grayscale + TrigRGB (Ours) & 34.3  & \textbf{75.24} & \textbf{97.5\%} \bigstrut[b]\\
\hline
\end{tabular}%

%% file: table/tokenreduce.tex
\begin{tabular}{lccccc}
\hline
Model & Gray Tokens/Frame & RGB(\%) & All   & \#VisualToken & $\Delta$ \bigstrut\\
\hline
\hline
Full RGB & -     & 100   & 77.20 & 100.0\% & 100.0\% \bigstrut[t]\\
Gray  & 256   & 0     & 68.76 & 100.0\% & 89.1\% \\
Gray  & 64    & 0     & 62.08 & 25.0\% & 80.4\% \bigstrut[b]\\
\hline
Gray + TrigRGB & 256   & 8.1   & 73.68 & 100.0\% & 95.4\% \bigstrut[t]\\
Gray + TrigRGB & 64    & 8.1   & 67.08 & 25.0\% & 86.9\% \\
\rowcolor[rgb]{ .984,  .886,  .835} Gray + TrigRGB + DT & 64    & 8.1   & 70.72 & \textbf{31.1\%} & 91.6\% \bigstrut[b]\\
\hline
Gray + TrigRGB & 256   & 34.3  & 76.32 & 100.0\% & 98.9\% \bigstrut[t]\\
Gray + TrigRGB & 64    & 34.3  & 68.96 & 25.0\% & 89.3\% \\
\rowcolor[rgb]{ .984,  .886,  .835} Gray + TrigRGB + DT & 64    & 34.3  & 75.24 & \textbf{50.7\%} & 97.5\% \bigstrut[b]\\
\hline
\end{tabular}%

%% file: sec/5_conclusion.tex
\section{Conclusion}
We introduced \method{}, a \textit{grayscale-always, color-on-demand} framework for efficient always-on streaming video sensing with MLLMs. 
By exploiting the natural redundancy of video through the similarity structure of a continuous grayscale stream, our training-free strategy triggers color acquisition only when informative. This design 
operates in a single pass with low latency, and achieves comparable performance using only a small fraction of RGB frames. 
We hope it inspires future research on always-on sensing.

%% file: sec/X_suppl.tex
\clearpage
\setcounter{page}{1}
\maketitlesupplementary

This supplementary material provides: 
(1) more feasibility discussion including power efficiency and latency (Sec.~\ref{sec:feasibility}), 
(2) additional experiments on Qwen3-VL~\cite{yang2025qwen3} validating the generalizability of \method{} (Sec.~\ref{sec:qwen3_experiments}), 
(3) more qualitative visualizations of our proposed method (Sec.~\ref{sec:visualizations}), 
and 
(4) discussion of limitations and future work (Sec.~\ref{sec:limitations}).

\section{Feasibility Analyses}
\label{sec:feasibility}
A key question for deploying \method{} on real wearable 
devices is whether the \emph{grayscale-always, 
color-on-demand} paradigm translates into tangible 
benefits under realistic hardware constraints. 
In this section, we analyze two critical aspects: 
power efficiency (Sec.~\ref{subsec:power}), where we 
quantify the energy gap between always-on grayscale and 
on-demand RGB capture on commercial components, and 
latency (Sec.~\ref{subsec:latency}), where we verify that 
the trigger pipeline operates within real-time budgets.

\subsection{Power Efficiency}
\label{subsec:power}
We position \method{abbr} as a \textbf{prototype} feasibility study for constrained wearables. Guided by industry norms~\cite{qualcomm_ar2_2022,qualcomm_ar2_platform,meta2025rayban}, we adopt a \textbf{split-compute} design: the wearable performs minimal capture and transmission, while heavy computation (\eg,\,trigger, VLM, Q\&A) runs on a companion server/cloud. This hardware-grounded assumption directs our analysis to the two dominant wearable energy costs:

\textbf{1) Capture (up to 90$\times$ Gap):} A dedicated always-on grayscale sensor operates at \textbf{$<$2\,mW} (\eg, Himax HM01B0~\cite{himax_hm01b0}), whereas commercial RGB sensors used in mainstream devices can be far higher with additional color-ISP processing (\eg, Sony IMX681~\cite{sony_imx681}: \textbf{182\,mW}), yielding up to a \textcolor{red}{\uline{\textbf{$\sim$90$\times$}}} acquisition efficiency gap.

\textbf{2) Transmission ($\sim$3$\times$ Gap):} For Bluetooth Low Energy (BLE)-class radios (active power magnitude $\sim$10--20\,mW~\cite{nordic_nrf52840}), RGB inherently carries 3$\times$ more raw data (3 channels \vs\,1) than grayscale. This implies a 
larger 
data payload, which proportionally extends radio-on airtime, resulting in higher energy.

\noindent\uline{\textbf{Battery Life:}} On a commercial device like the Meta Ray-Ban~\cite{raybanmeta_markings} (154\,mAh battery), continuously keeping RGB recording active would drain the battery in \textbf{$<$1\,hour} due to the high hardware baseline~\cite{meta2025rayban}. In contrast, utilizing grayscale sensing lowers the power floor to enable toward all-day operation (approx. \textbf{$>$10\,hours}), making always-on video understanding physically viable. 

We note that the $\sim$90$\times$ gap compares sensors at 
different resolutions (12\,MP RGB \vs\, 0.1\,MP grayscale); 
this reflects realistic hardware pairings, as manufacturers 
require high-resolution RGB for capture quality while 
favoring ultra-low-power grayscale for always-on monitoring. 
Even at matched resolution, the asymmetry persists, as RGB 
additionally requires color filtering, ISP processing, and 
3$\times$ the transmission payload.

\begin{table}[t] 
\centering
\caption{Performance comparison on Video-MME~\cite{fu2025videomme}.}
\resizebox{\linewidth}{!}{%
\input{table/videomme_qwen3}
}
\label{tab:videomme_qwen3}
\end{table}

\subsection{Latency}
\label{subsec:latency}

Our test on an Intel Xeon shows CLIP+QP takes 275 ms/frame (QP: 5.5 ms). Further validating edge feasibility, switching to MobileCLIP2-S0~\cite{faghri2025mobileclip2} still yields avg 70.47 at 8.1\% RGB, which achieves only \textbf{1.5\,ms}/frame visual encoding on iPhone12PM~\cite{faghri2025mobileclip2}.

For the remaining, we give a compact estimate from public references: a 224$\times$224 8-bit gray frame is $\sim$50\,KB, yielding $\sim$0.3\,s over BLE ($\sim$1.4\,Mbps under LE 2M~\cite{bluetooth5_spec}); camera/ISP wake and image capture are hardware-dependent but typically \textbf{tens of ms}. Under low-rate always-on sampling (\eg,\,1\,fps), this supports practical online color-on-demand decisions.

\begin{table*}[ht]
\centering
\caption{Performance comparison on StreamingBench~\cite{lin2024streamingbench} focusing on \textit{Real-Time Visual Understanding} tasks. Real-Time Visual Understanding encompasses Object Perception (OP), Causal Reasoning (CR), Clips Summarization (CS), Attribute Perception (ATP), Event Understanding (EU), Text-Rich Understanding (TR), Prospective Reasoning (PR), Spatial Understanding (SU), Action Perception (ACP), and Counting (CT). ``{RGB(\%)}'' represents the percentage of color frames remaining after triggering, where $100$ indicates full color, $0$ means all grayscale, and $\downarrow91.9\%$ signifies a 91.9\% reduction in the number of color frames. 
}
\resizebox{\linewidth}{!}{%
\input{table/streamingbench_qwen3}
}
\label{tab:streamingbench_qwen3}
\end{table*}

\begin{table*}[ht]
\centering
\caption{Results on OVO-Bench~\cite{niu2025ovo} comprising three categories: i) \textit{Real-Time Visual Perception} (OCR: Optical Character Recognition, ACR: Action Recognition, ATR: Attribute Recognition, STU: Spatial Understanding, FPD: Future Prediction, OJR: Object Recognition), ii) \textit{Backward Tracing} (EPM: Episodic Memory, ASI: Action Sequence Identification, HLD: Hallucination Detection), and iii) \textit{Forward Active Responding} (REC: Repetition Event Count, SSR: Sequential Steps Recognition, CRR: Clues Reveal Responding).}    
\resizebox{\linewidth}{!}{%
\input{table/ovobench_qwen3.tex}
}
\label{tab:ovobench_qwen3}
\end{table*}

\section{More results on Qwen3-VL}
\label{sec:qwen3_experiments}

In this section, we conduct experiments by applying Qwen3-VL-8B-Instruct~\cite{yang2025qwen3} as the MLLM backbone to further validate the generalizability of our proposed \method{} across multiple mainstream open-source MLLMs. These results demonstrate that our framework achieves consistent improvements across different model architectures and training paradigms.

\noindent\textbf{Implementation for Qwen3-VL.}
Unlike InternVL-3.5~\cite{wang2025internvl35}, which processes frames independently, Qwen3-VL employs temporal window attention in its visual encoder, where frames attend to each other within temporal windows. This architectural difference prevents us from using asymmetric token budgets for grayscale and RGB frames, as the attention mechanism requires consistent spatial dimensions across the temporal sequence. To address this constraint, we apply bicubic upsampling to resize grayscale frames from $224\times224$ to $448\times448$, matching the default resolution of RGB frames in Qwen3-VL's processing pipeline. While this removes the token-level computational savings from our Dynamic Token Router, the primary benefit of reduced RGB frame acquisition and sensor power consumption remains intact.

\noindent\textbf{Performance analysis.}
Table~\ref{tab:streamingbench_qwen3} and Table~\ref{tab:ovobench_qwen3} present results on StreamingBench and OVO-Bench, respectively. On both benchmarks, \method{} demonstrates consistent effectiveness when applied to Qwen3-VL. With minimal RGB frame usage, our approach substantially outperforms the grayscale-only baseline while recovering most of the full RGB baseline performance. When allowing moderate RGB frame usage, our model closely matches the full RGB performance while using only a fraction of chromatic frames. These trends are consistent across both StreamingBench and OVO-Bench, validating the robustness of our framework across different evaluation protocols.

Notably, due to the temporal attention mechanism in Qwen3-VL's visual encoder, where frames influence each other's representations, we observe minor fluctuations in certain fine-grained spatial reasoning subtasks. For instance, on OVO-Bench's STU (Spatial Understanding) subtask, the performance variance is slightly larger compared to InternVL-based models. This is expected behavior, as the temporal cross-frame attention can propagate visual information differently when grayscale and RGB frames are interleaved, occasionally affecting spatial localization precision. Despite these task-specific variations, the overall performance across comprehensive benchmarks remains strong and consistent.

Table~\ref{tab:videomme_qwen3} presents results on Video-MME, which evaluates video understanding across short, medium, and long duration categories. \method{} demonstrates consistent improvements over the grayscale-only baseline across all video lengths. With moderate RGB frame usage, our model achieves performance comparable to the full RGB baseline while substantially reducing chromatic frame acquisition. The results on Video-MME further validate that our adaptive triggering mechanism effectively identifies critical moments requiring color information across videos of varying temporal complexity. Notably, the performance gains are consistent across all duration categories, indicating that \method{} maintains robust temporal coherence and semantic understanding regardless of video length. These findings complement our observations on StreamingBench and OVO-Bench, collectively demonstrating that our framework generalizes effectively across diverse video understanding tasks and temporal scales when applied to Qwen3-VL architecture.

These results confirm that \method{} generalizes effectively across different MLLM architectures, maintaining substantial RGB frame reduction while preserving strong video understanding capabilities, even when architectural constraints prevent full exploitation of our dynamic tokenization strategy.

\begin{figure*}[t]
\centering
\includegraphics[width=1\linewidth]{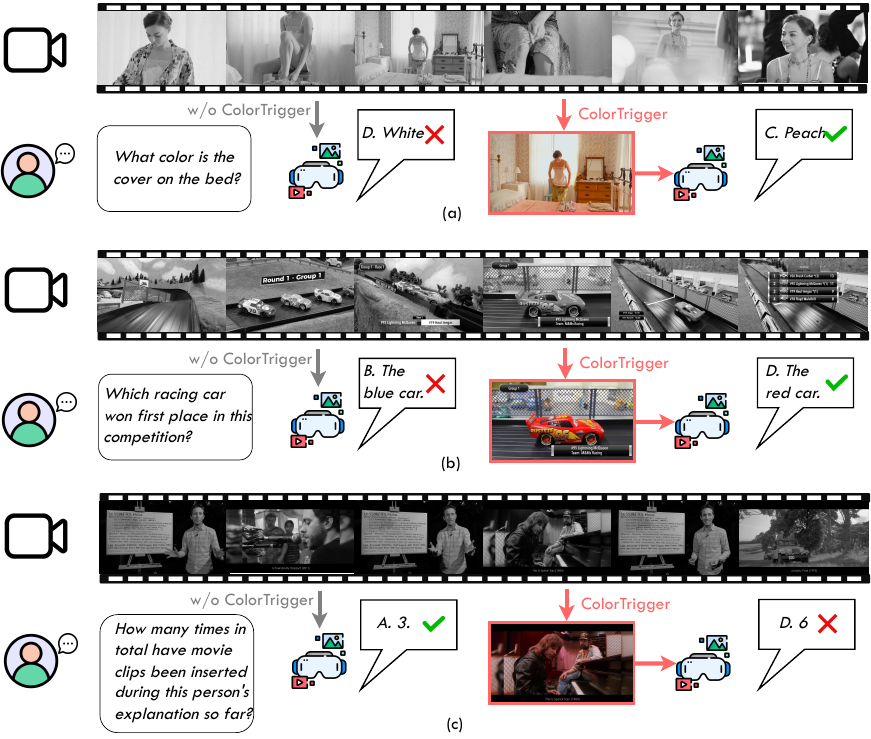}
\caption{
\textbf{More qualitative examples of \protect\method{}.} 
Our causal trigger selectively injects RGB frames (highlighted in red boxes) when 
chromatic information is critical for answering queries. 
\textbf{(a-b)} Success cases: questions requiring color discrimination (bed cover color, 
car identification) benefit from triggered RGB frames, correcting errors made without 
ColorTrigger. 
\textbf{(c)} Failure case: hard insertion of RGB frames disrupts temporal continuity, 
causing the MLLM to miscount movie clip insertions due to perceived semantic discontinuities 
(discussed in Sec.~\ref{sec:limitations}).
}
\label{fig:qualitative}
\label{fig:visualization_suppl} 
\end{figure*}

\section{More visualization}
\label{sec:visualizations}

Figure~\ref{fig:visualization_suppl} shows representative examples. 
Cases (a-b) demonstrate successful color-based reasoning: without ColorTrigger, the MLLM 
hallucinates incorrect answers (``white'' bed cover, ``blue'' car) due to missing chromatic 
cues; with our trigger, RGB frames are selectively injected at critical moments, enabling 
correct responses. 
However, case (c) exposes a failure mode: the question asks to count movie clip insertions, 
which requires detecting temporal transitions. 
Our hard frame insertion introduces artificial discontinuities that the model misinterprets as additional boundaries, 
inflating the count from 3 to 6. 
This limitation is further discussed in Sec.~\ref{sec:limitations}.

\section{Limitations and Future Work}
\label{sec:limitations}

While \method{} achieves strong efficiency-accuracy trade-offs, our design introduces 
temporal discontinuities in the visual stream due to the gap between grayscale and RGB frames. 
The abrupt transitions in \emph{spatial resolution} ($224 \times 224$ vs. $448 \times 448$), 
\emph{chromatic content} (L-channel only \vs\,  full RGB), and \emph{token density} 
($T_g=64$ vs. $T_c=256$) create perceptual ``jumps'' that may disrupt the temporal 
coherence expected by the frozen MLLM decoder.
This mismatch is particularly salient for tasks requiring fine-grained temporal 
reasoning, such as tracking gradual motion or understanding smooth action sequences, 
where the hard switching mechanism can partially offset 
efficiency and performance gains.

Despite these artifacts, our experiments show that the impact on 
aggregate accuracy remains modest, suggesting that many video understanding tasks exhibit 
sufficient robustness to modality transitions. 
Nonetheless, an important direction for future work is to explore alternative 
integration strategies that preserve temporal coherence without requiring abrupt frame 
insertions, such as learning smooth feature-space transitions or designing hybrid 
representations that naturally bridge grayscale and RGB modalities while maintaining 
our strict causality and training-free constraints.

Our trigger is designed to suppress redundant color 
captures in temporally stable scenes. In high-mobility 
scenarios (\eg, walking through a busy street), frequent 
scene changes may lead to near-continuous RGB activation, 
reducing the power savings. 
A natural extension is to 
incorporate \emph{user-intent awareness}: for proactive 
agents, a lightweight module could estimate whether the 
current scene is task-relevant before triggering RGB; 
for passive agents, a compressed visual memory could 
defer detailed processing until the user issues a query. 

We leave these directions to future work.

%% file: table/videomme_qwen3.tex
\begin{tabular}{lcccccc}
\hline
\multirow{2}[4]{*}{Method} & \multirow{2}[4]{*}{\#Frames} & \multirow{2}[4]{*}{RGB (\%)} & \multicolumn{4}{c}{Video-MME} \bigstrut\\
\cline{4-7}      &       &       & short & medium & long  & overall \bigstrut\\
\hline
\hline
VideoChat2-7B~\cite{li2024mvbench} & 16    & 100   & 48.3  & 37.0  & 33.2  & 39.5 \bigstrut[t]\\
LongVA-7B~\cite{zhang2024longva} & 128   & 100   & 61.1  & 50.4  & 46.2  & 52.6 \\
Kangaroo-7B~\cite{kangaroogroup} & 64    & 100   & 66.1  & 55.3  & 46.6  & 56.0 \\
Video-CCAM-14B~\cite{fei2024video} & 96    & 100   & 62.2  & 50.6  & 46.7  & 53.2 \\
VideoXL-7B~\cite{shu2025videoxl} & 128   & 100   & 64.0  & 53.2  & 49.2  & 55.5 \\
Dispider-7B~\cite{qian2025dispider} & 1 fps & 100   & -     & -     & -     & 57.2 \\
VideoChat-Online-4B~\cite{huang2024online} & 2 fps & 100   & -     & -     & 47.1  & 54.4 \\
TimeChat-Online-7B~\cite{yao2025timechatonline} & 1 fps & 100   & -     & -     & 48.4  & 62.4 \bigstrut[b]\\
\hline
InternVL-3.5-8B & 128   & 100   & 76.7  & 65.3  & 54.7  & 65.6 \bigstrut[t]\\
\rowcolor[rgb]{ .91,  .91,  .91} InternVL-3.5-8B grayscale & 128   & 0     & 63.4  & 57.9  & 50.6  & 57.3 \\
\rowcolor[rgb]{ .984,  .886,  .835} InternVL-3.5 + \method{} (Ours, 8B) & 128   & 9.1 (\downpct{90.9}) & 71.9  & 63.0  & 53.6  & 62.8 (\uppct{5.5}) \\
\rowcolor[rgb]{ .969,  .78,  .675} InternVL-3.5 + \method{} (Ours, 8B) & 128   & 37.6 (\downpct{62.4}) & 76.8  & 66.3  & 55.1  & 66.1 (\uppct{8.8}) \bigstrut[b]\\
\hline
Qwen3-VL-8B & 128   & 100   & 79.0  & 68.1  & 58.1  & 68.4 \bigstrut[t]\\
\rowcolor[rgb]{ .91,  .91,  .91} Qwen3-VL-8B grayscale & 128   & 0     & 70.8  & 64.6  & 56.2  & 63.9 \\
\rowcolor[rgb]{ .855,  .949,  .816} Qwen3 + \method{} (Ours, 8B) & 128   & 9.1 (\downpct{90.9}) & 74.8  & 66.6  & 58.2  & 66.5 (\uppct{2.6}) \\
\rowcolor[rgb]{ .71,  .902,  .635} Qwen3 + \method{} (Ours, 8B) & 128   & 37.6 (\downpct{62.4}) & 77.1  & 66.6  & 59.0  & 67.6 (\uppct{3.7}) \bigstrut[b]\\
\hline
\end{tabular}%

%% file: table/streamingbench_qwen3.tex
\begin{tabular}{lccccccccccccc}
\hline
Model & \#Frames & RGB (\%) & OP    & CR    & CS    & ATP   & EU    & TR    & PR    & SU    & ACP   & CT    & All \bigstrut\\
\hline
\hline
Human & -     & -     & 89.47 & 92.00 & 93.60 & 91.47 & 95.65 & 92.52 & 88.00 & 88.75 & 89.74 & 91.30 & 91.46 \bigstrut\\
\hline
\multicolumn{14}{c}{\textbf{Proprietary MLLMs}} \bigstrut\\
\hline
Gemini 1.5 pro~\cite{team2024gemini} & 1 fps & 100   & 79.02 & 80.47 & 83.54 & 79.67 & 80.00 & 84.74 & 77.78 & 64.23 & 71.95 & 48.70 & 75.69 \bigstrut[t]\\
GPT-4o~\cite{hurst2024gpt} & 64    & 100   & 77.11 & 80.47 & 83.91 & 76.47 & 70.19 & 83.80 & 66.67 & 62.19 & 69.12 & 49.22 & 73.28 \\
Claude 3.5 Sonnet~\cite{claude3_5} & 20    & 100   & 80.49 & 77.34 & 82.02 & 81.73 & 72.33 & 75.39 & 61.11 & 61.79 & 69.32 & 43.09 & 72.44 \bigstrut[b]\\
\hline
\multicolumn{14}{c}{\textbf{Open-Source Video MLLMs }} \bigstrut\\
\hline
LLaVA-OneVision-7B~\cite{li2024llavaonevision} & 32    & 100   & 80.38 & 74.22 & 76.03 & 80.72 & 72.67 & 71.65 & 67.59 & 65.45 & 65.72 & 45.08 & 71.12 \bigstrut[t]\\
Video-LLaMA2-7B~\cite{cheng2024videollama2} & 32    & 100   & 55.86 & 55.47 & 57.41 & 58.17 & 52.80 & 43.61 & 39.81 & 42.68 & 45.61 & 35.23 & 49.52 \\
Qwen2.5-VL-7B~\cite{bai2025qwen25vl} & 1 fps & 100   & 78.32 & 80.47 & 78.86 & 80.45 & 76.73 & 78.50 & 79.63 & 63.41 & 66.19 & 53.19 & 73.68 \bigstrut[b]\\
\hline
\multicolumn{14}{c}{\textbf{Streaming MLLMs}} \bigstrut\\
\hline
Flash-VStream-7B~\cite{zhang2024flash} & -     & 100   & 25.89 & 43.57 & 24.91 & 23.87 & 27.33 & 13.08 & 18.52 & 25.20 & 23.87 & 48.70 & 23.23 \bigstrut[t]\\
VideoLLM-online-8B~\cite{chen2024videollmonline} & 2 fps & 100   & 39.07 & 40.06 & 34.49 & 31.05 & 45.96 & 32.40 & 31.48 & 34.16 & 42.49 & 27.89 & 35.99 \\
Dispider-7B~\cite{qian2025dispider} & 1 fps & 100   & 74.92 & 75.53 & 74.10 & 73.08 & 74.44 & 59.92 & 76.14 & 62.91 & 62.16 & 45.80 & 67.63 \\
TimeChat-Online-7B~\cite{yao2025timechatonline} & 1 fps & 100   & 80.22 & 82.03 & 79.50 & 83.33 & 76.10 & 78.50 & 78.70 & 64.63 & 69.60 & 57.98 & 75.36 \bigstrut[b]\\
\hline
InternVL-3.5-8B & 128   & 100   & 83.47 & 82.03 & 82.65 & 84.62 & 75.47 & 80.06 & 81.48 & 67.89 & 70.45 & 59.04 & 77.20 \bigstrut[t]\\
\rowcolor[rgb]{ .91,  .91,  .91} InternVL-3.5-8B grayscale & 128   & 0     & 60.98 & 78.91 & 60.88 & 55.45 & 65.41 & 63.24 & 75.00 & 60.16 & 62.22 & 55.85 & 62.08 \\
\rowcolor[rgb]{ .984,  .886,  .835} InternVL-3.5 + \method{} (Ours, 8B) & 128   & 8.1 (\downpct{91.9}) & 75.07 & 77.34 & 72.56 & 75.64 & 71.70 & 70.40 & 76.85 & 65.04 & 66.48 & 57.98 & 70.72 (\uppct{8.64}) \\
\rowcolor[rgb]{ .969,  .78,  .675} InternVL-3.5 + \method{} (Ours, 8B) & 128   & 34.3 (\downpct{65.7}) & 81.57 & 78.91 & 77.29 & 84.94 & 76.10 & 77.57 & 81.48 & 67.48 & 67.61 & 56.91 & 75.24 (\uppct{13.16}) \bigstrut[b]\\
\hline
Qwen3-VL-8B & 128   & 100   & 76.78 & 75.78 & 79.81 & 80.45 & 71.70 & 77.50 & 77.78 & 68.16 & 65.62 & 56.38 & 73.43 \bigstrut[t]\\
\rowcolor[rgb]{ .91,  .91,  .91} Qwen3-VL-8B grayscale & 128   & 0     & 65.57 & 70.31 & 76.34 & 56.41 & 65.41 & 70.31 & 76.85 & 62.45 & 63.64 & 53.19 & 65.61 \\
\rowcolor[rgb]{ .855,  .949,  .816} Qwen3 + \method{} (Ours, 8B) & 128   & 8.1 (\downpct{91.9}) & 68.03 & 70.31 & 76.97 & 70.83 & 67.92 & 73.44 & 75.93 & 61.63 & 63.64 & 53.19 & 68.30 (\uppct{2.69}) \\
\rowcolor[rgb]{ .71,  .902,  .635} Qwen3 + \method{} (Ours, 8B) & 128   & 34.3 (\downpct{65.7}) & 75.41 & 74.22 & 78.86 & 79.81 & 75.47 & 77.19 & 77.78 & 62.45 & 64.77 & 54.26 & 72.30 (\uppct{6.69}) \bigstrut[b]\\
\hline
\end{tabular}%

%% file: table/ovobench_qwen3.tex
\begin{tabular}{lcccccccccccccccccc}
\hline
\multirow{2}[4]{*}{Model} & \multirow{2}[4]{*}{\#Frames} & \multirow{2}[4]{*}{RGB (\%)} & \multicolumn{7}{c}{Real-Time Visual Perception}       & \multicolumn{4}{c}{Backward Tracing} & \multicolumn{4}{c}{Forward Active Responding} & Overall \bigstrut\\
\cline{4-18}      &       &       & OCR   & ACR   & ATR   & STU   & FPD   & OJR   & Avg.  & EPM   & ASI   & HLD   & Avg.  & REC   & SSR   & CRR   & Avg.  & Avg. \bigstrut\\
\hline
\hline
Human Agents & -     & -     & 94.0  & 92.6  & 94.8  & 92.7  & 91.1  & 94.0  & 93.2  & 92.6  & 93.0  & 91.4  & 92.3  & 95.5  & 89.7  & 93.6  & 92.9  & 92.8 \bigstrut\\
\hline
\multicolumn{19}{c}{\textbf{Proprietary MLLMs}} \bigstrut\\
\hline
Gemini 1.5 Pro~\cite{team2024gemini} & 1 fps & 100   & 87.3  & 67.0  & 80.2  & 54.5  & 68.3  & 67.4  & 70.8  & 68.6  & 75.7  & 52.7  & 62.3  & 35.5  & 74.2  & 61.7  & 57.2  & 65.3 \bigstrut[t]\\
GPT-4o~\cite{hurst2024gpt} & 64    & 100   & 69.1  & 65.1  & 65.5  & 50.0  & 68.3  & 63.7  & 63.6  & 49.8  & 71.0  & 55.4  & 58.7  & 27.6  & 73.2  & 59.4  & 53.4  & 58.6 \bigstrut[b]\\
\hline
\multicolumn{19}{c}{\textbf{Open-Source Video MLLMs }} \bigstrut\\
\hline
LLaVA-NeXT-Video-7B~\cite{llavanextvideo} & 64    & 100   & 69.8  & 59.6  & 66.4  & 50.6  & 72.3  & 61.4  & 63.3  & 51.2  & 64.2  & 9.7   & 41.7  & 34.1  & 67.6  & 60.8  & 54.2  & 53.1 \bigstrut[t]\\
LLaVA-OneVision-7B~\cite{li2024llavaonevision} & 64    & 100   & 67.1  & 58.7  & 69.8  & 49.4  & 71.3  & 60.3  & 62.8  & 52.5  & 58.8  & 23.7  & 45.0  & 24.8  & 66.9  & 60.8  & 50.9  & 52.9 \\
Qwen2-VL-7B~\cite{wang2024qwen2vl} & 64    & 100   & 69.1  & 53.2  & 63.8  & 50.6  & 66.3  & 60.9  & 60.7  & 44.4  & 66.9  & 34.4  & 48.6  & 30.1  & 65.7  & 50.8  & 48.9  & 52.7 \\
LongVU-7B~\cite{shen2024longvu} & 1 fps & 100   & 55.7  & 49.5  & 59.5  & 48.3  & 68.3  & 63.0  & 57.4  & 43.1  & 66.2  & 9.1   & 39.5  & 16.6  & 69.0  & 60.0  & 48.5  & 48.5 \bigstrut[b]\\
\hline
\multicolumn{19}{c}{\textbf{Streaming Video-LLMs}} \bigstrut\\
\hline
Flash-VStream-7B~\cite{zhang2024flash} & 1 fps & 100   & 25.5  & 32.1  & 29.3  & 33.7  & 29.7  & 28.8  & 29.9  & 36.4  & 33.8  & 5.9   & 25.4  & 5.4   & 67.3  & 60.0  & 44.2  & 33.2 \bigstrut[t]\\
VideoLLM-online-8B~\cite{chen2024videollmonline} & 2 fps & 100   & 8.1   & 23.9  & 12.1  & 14.0  & 45.5  & 21.2  & 20.8  & 22.2  & 18.8  & 12.2  & 17.7  & -     & -     & -     & -     & - \\
TimeChat-Online-7B~\cite{yao2025timechatonline} & 1 fps & 100   & 75.2  & 46.8  & 70.7  & 47.8  & 69.3  & 61.4  & 61.9  & 55.9  & 59.5  & 9.7   & 41.7  & 31.6  & 38.5  & 40.0  & 36.7  & 46.7 \\
Qwen2.5-VL-7B~\cite{bai2025qwen25vl} & 1 fps & 100   & 73.8  & 56.0  & 68.1  & 46.6  & 71.3  & 60.3  & 62.7  & 48.2  & 64.9  & 26.9  & 46.6  & 36.2  & 41.8  & 47.1  & 41.7  & 50.3 \bigstrut[b]\\
\hline
InternVL-3.5-8B & 128   & 100   & 77.9  & 64.2  & 75.9  & 56.7  & 75.3  & 70.7  & 70.1  & 56.6  & 66.2  & 34.4  & 52.4  & 48.2  & 62.7  & 41.3  & 50.7  & 57.7 \bigstrut[t]\\
\rowcolor[rgb]{ .91,  .91,  .91} InternVL-3.5-8B grayscale & 128   & 0     & 61.1  & 46.8  & 53.5  & 47.8  & 56.4  & 57.1  & 53.8  & 46.1  & 54.7  & 37.1  & 46.0  & 35.3  & 54.3  & 42.1  & 43.9  & 47.9 \\
\rowcolor[rgb]{ .984,  .886,  .835} \method{} (Ours, 8B) & 128   & 7.1 (\downpct{92.9}) & 65.1  & 52.3  & 61.2  & 50.6  & 66.3  & 63.6  & 59.9 (\uppct{6.1}) & 50.5  & 58.1  & 32.3  & 47.0  & 32.5  & 59.3  & 41.3  & 44.3  & 50.4 (\uppct{2.5}) \\
\rowcolor[rgb]{ .969,  .78,  .675} \method{} (Ours, 8B) & 128   & 33.1  (\downpct{66.9}) & 77.9  & 56.0  & 69.8  & 55.1  & 67.3  & 65.2  & 65.2 (\uppct{11.4}) & 50.2  & 62.8  & 30.7  & 41.3  & 32.7  & 59.4  & 40.8  & 44.3  & 52.5 (\uppct{4.6}) \bigstrut[b]\\
\hline
Qwen3-VL-8B & 128   & 100   & 77.9  & 60.6  & 71.6  & 56.7  & 68.0  & 60.3  & 65.8  & 52.9  & 71.0  & 18.8  & 47.5  & 44.6  & 66.9  & 48.3  & 53.3  & 55.6 \bigstrut[t]\\
\rowcolor[rgb]{ .91,  .91,  .91} Qwen3-VL-8B grayscale & 128   & 0     & 68.5  & 55.1  & 51.7  & 53.9  & 71.0  & 49.5  & 58.3  & 48.8  & 64.2  & 29.0  & 47.4  & 43.6  & 61.9  & 45.4  & 50.3  & 52.0 \\
\rowcolor[rgb]{ .855,  .949,  .816} Qwen3 + \method{} (Ours, 8B) & 128   & 7.1 (\downpct{92.9}) & 71.8  & 59.6  & 63.8  & 55.1  & 71.0  & 55.4  & 62.8 (\uppct{4.5}) & 51.2  & 66.2  & 26.3  & 47.9  & 47.5  & 62.3  & 45.0  & 51.6  & 54.1 (\uppct{2.1}) \\
\rowcolor[rgb]{ .71,  .902,  .635} Qwen3 + \method{} (Ours, 8B) & 128   & 33.1  (\downpct{66.9}) & 75.2  & 62.4  & 71.6  & 52.8  & 72.0  & 57.1  & 65.2 (\uppct{6.9}) & 51.9  & 68.2  & 18.3  & 46.1  & 43.3  & 66.0  & 47.9  & 52.4  & 54.6 (\uppct{2.6}) \bigstrut[b]\\
\hline
\end{tabular}%